\documentclass[letterpaper]{article}
\usepackage{aaai2027}
\nocopyright
\usepackage{caption}
\usepackage[hyphens]{url}
\usepackage{graphicx}
\usepackage{natbib}
\usepackage{booktabs}
\usepackage{array,tabularx}
\usepackage[table]{xcolor}
\usepackage{amsmath,amssymb,amsfonts}
\allowdisplaybreaks[1]

\newcounter{seniorfn}
\newcommand{\seniorcontrib}{%
  \ifnum\value{seniorfn}=0%
    \footnote{These authors share senior authorship.}%
    \setcounter{seniorfn}{\value{footnote}}%
  \else%
    \footnotemark[\value{seniorfn}]%
  \fi%
}

\title{Learning to Coordinate Symbolic Tools: LLM Agents for Verified Sum-of-Squares Certificates}

\author{
Bohan Chen\equalcontrib\corresponding\textsuperscript{\rm 1},
Shivam N.\ Patel\equalcontrib\footnote{This work was completed while S. N. Patel was a research visitor at the California Institute of Technology, prior to his affiliation with OpenAI.}\textsuperscript{\rm 1},
Richard Hoffmann\textsuperscript{\rm 1},\\
Sam Looi\seniorcontrib\textsuperscript{\rm 1},
Tony Yue Yu\seniorcontrib\textsuperscript{\rm 1}
}
\affiliations{
\textsuperscript{\rm 1}California Institute of Technology\\
\textsuperscript{\rm 2}OpenAI
}

\date{}

\begin{document}

\maketitle

\begin{abstract}
Tool calling allows large language models (LLMs) to invoke external computation during problem solving, a useful capability in various fields including AI for mathematics. We study this setting through weighted sum-of-squares (SOS) decomposition, a machine-checkable route to proving polynomial nonnegativity and hence polynomial inequalities. A candidate decomposition can be checked exactly, but finding one requires choosing among non-unique regroupings and coordinating multiple symbolic transformations. We develop an agent that combines algebraic task training, symbolic tools, and verifier-grounded optimization for this task. Rather than training only on the composite SOS task, we construct 1.35 million synthetic examples covering eight supporting polynomial tasks together with weighted-SOS decomposition. We first apply supervised fine-tuning (SFT) to direct algebra problems and simulated symbolic traces, and then use Group Relative Policy Optimization (GRPO) with task-specific symbolic rewards. The SFT corpus contains no native tool-calling messages; at evaluation, the agent uses native SymPy calls for expansion, collection, reordering, and factorization. Every final SOS answer is checked by exact expansion and coefficient comparison. On held-out, same-generator synthetic problems, the full SFT+GRPO+tools system is the strongest of four evaluated configurations, reaching 78.96\% verified success on weighted SOS, compared with 44.73\% for the base model with the same tools, and 91.75\% macro accuracy across nine polynomial tasks. Within this controlled setting, our work provides a case study of combining domain-specific skill training, executable tools, and verifier feedback, and may inform the design of tool-calling agents in other domains with exactly checkable outputs.
\end{abstract}

\section{Introduction}

Exact tools can make an individual operation reliable without making an agent's overall strategy correct. A tool-using language model must still decide which operation is useful, construct its arguments, interpret the observation, recover from an unhelpful result, and know when to stop \cite{Yao2023ReAct, Li2023APIBank}. These decisions are especially visible in tasks whose outputs are easy to verify but difficult to search for \cite{yang2023intercodestandardizingbenchmarkinginteractive}. 
This leads to our central question:
\emph{Can algebra-grounded post-training enable an LLM to coordinate exact symbolic operations during multi-step symbolic search, rather than merely execute isolated algebraic transformations?}

Polynomials are a foundational language for optimization and algebraic geometry. Recent developments around the Jacobian conjecture make their local--global subtlety especially vivid \citep{Ramos2026JacobianCounterexample, Tao2026JacobianCounterexample}. The mathematical task we study is different, but the episode underscores the value of exact polynomial reasoning and machine-checkable witnesses. We are well-motivated in our use of weighted SOS as a controlled setting in which discovering global structure requires search, while checking a proposed identity is exact and inexpensive.

A weighted sum-of-squares (SOS) certificate is an identity of the form
\begin{equation}
\label{eq:intro-sos}
f=\sum_{j=1}^{m}c_jp_j^2,\qquad c_j\in\mathbb{Q}_{>0}.
\end{equation}
Such an identity proves global polynomial nonnegativity and can be
checked exactly by expansion. Finding such an identity is nevertheless non-unique and strategic: expansion hides square structure, mixed terms admit several plausible groupings, and a locally correct transformation can leave an unusable remainder. Symbolic tools execute a proposed operation; they do not determine the global search path.

\begin{figure*}[t]
\centering
\includegraphics[width=\textwidth]{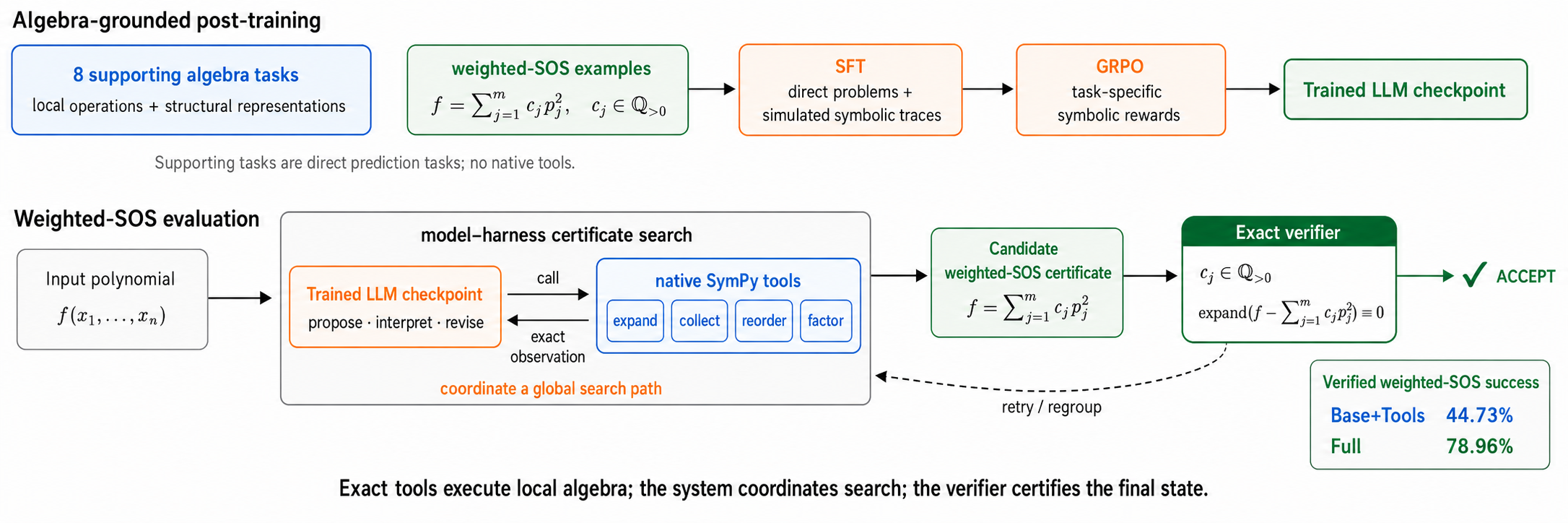}
\caption{Overview of the controlled study. Algebraic post-training builds a checkpoint from eight supporting tasks and weighted-SOS examples. During weighted-SOS evaluation, Base+Tools and Full receive the same native SymPy interface. The complete model--harness system coordinates exact local operations, while the terminal verifier accepts only a positive-rational weighted-SOS expression whose expansion matches the input polynomial exactly.}
\label{fig:overview}
\end{figure*}

Figure~\ref{fig:overview} summarizes the training and deployment boundary. Starting from \texttt{Phi-4-reasoning-plus} \citep{Abdin2025Phi4Reasoning}, we construct a nine-task synthetic curriculum: four local operations, four broader algebraic-structure tasks, and composite weighted-SOS search. We apply assistant-token QLoRA SFT \citep{Dettmers2023QLoRA} to direct problems and simulated symbolic traces, then continue with GRPO using task-specific symbolic rewards \citep{Shao2024DeepSeekMath}. The SFT corpus contains evaluated Python-like symbolic snippets but no native function-call messages. During weighted-SOS evaluation, the tool-enabled systems instead interact with four SymPy-backed functions \citep{Meurer2017SymPy}, and every terminal certificate is checked structurally and coefficient by coefficient.

On $90{,}000$ separately generated, same-generator test instances, the complete system reaches a $91.75\%$ macro-average across the nine polynomial tasks. On weighted SOS---the only native tool-calling task---the Full system reaches $78.96\%$ verified success, compared with $44.73\%$ for the same base backbone with the same tool interface. The experiment is deliberately controlled: it uses small synthetic polynomials, one backbone, and one run per training stage. Because the evaluated configurations do not include SFT+Tools without GRPO or GRPO without tools, the results compare complete systems rather than causally isolating individual components.

\paragraph{Contributions.}
\begin{itemize}
    \item \textbf{An exactly verifiable agent testbed.} We instantiate symbolic-tool coordination as weighted-SOS certificate search, supported by eight direct algebraic tasks, $1.35$ million generated examples before the validation split, and $90{,}000$ separately generated test problems.
    \item \textbf{An algebra-grounded post-training recipe.} The pipeline combines assistant-token QLoRA SFT on direct algebra and simulated symbolic traces, verifier-grounded GRPO, a native four-function SymPy interface for SOS, and exact terminal verification.
    \item \textbf{Complete-system evidence with calibrated scope.} We compare Base, SFT, Base+Tools, and Full on weighted SOS, report direct-task checkpoint performance and terminal failure categories, and distinguish the empirical findings from broader design hypotheses.
\end{itemize}

\section{Related Work}

\paragraph{Tool-using language-model agents.}
Toolformer learns where to insert API calls, while ReAct interleaves reasoning, actions, and observations \citep{Schick2023Toolformer,Yao2023ReAct}. API-Bank and ToolLLM broaden supervised tool learning to large API collections \citep{Li2023APIBank,Qin2024ToolLLM}. For mathematics, ToRA trains on interactive trajectories that alternate language reasoning with computation and symbolic solvers, while ReTool couples cold-start traces with outcome-driven reinforcement learning for strategic code use \citep{Gou2024ToRA,Feng2026ReTool}. Evaluation has likewise moved beyond isolated call syntax: MINT studies multi-turn interaction with tool and language feedback, BFCL includes abstention and stateful multi-step use, and $\tau$-bench scores the final environment state and repeated-trial reliability \citep{Wang2024MINT,Patil2025BFCL,Yao2025TauBench}. Multi-turn studies further show that premature assumptions can make later recovery unreliable \citep{Laban2025LostConversation}. Our environment is narrower but more exact: the tools perform deterministic algebra, and the terminal state is a coefficient identity. CRITIC uses tools mainly to critique and revise an answer \citep{Gou2024CRITIC}; here exact operations are used constructively throughout certificate search.

\paragraph{Verifier-guided mathematical reasoning.}
PAL and Program of Thoughts delegate computation to executable programs, and ToRA makes this interaction iterative rather than one-shot \citep{Gao2023PAL,Chen2023ProgramOfThoughts,Gou2024ToRA}. Outcome and process verifiers provide complementary supervision for multi-step mathematics \citep{Cobbe2021GSM8K,Lightman2024VerifyStep}. DeepSeekMath and ReTool use verifiable outcomes to optimize mathematical reasoning, while LeanDojo, DeepSeek-Prover, and AlphaProof ground search in proof-assistant states checked by a formal kernel \citep{Shao2024DeepSeekMath,Yang2023LeanDojo,Xin2025DeepSeekProverV15,Ren2025DeepSeekProverV2,Hubert2025AlphaProof}. Exact behavior can nevertheless remain fragile under controlled
numerical or contextual perturbations, or as combinatorial complexity
increases \citep{Mirzadeh2024GSMSymbolic,Shojaee2025IllusionThinking}. Although our verifier is less expressive than Lean, it is cheaper and deterministic by checking the submitted algebraic object rather than a natural-language rationale.

\paragraph{Polynomial certificates and neuro-symbolic proving.}
SOS and Positivstellensatz certificates connect polynomial nonnegativity to algebraic identities and semidefinite optimization \citep{Stengle1974,Parrilo2003SDPRelaxations,Lasserre2001GlobalOptimization}. Learning has been used to guide dynamic polynomial proofs and symbolic inequality search: AIPS learns search guidance, LIPS assigns scaling to symbolic routines and rewriting/ranking to an LLM, and IneqSearch searches for combinations of nonnegative components \citep{Fawzi2019DynamicPolynomialProofs,Wei2024AIPS,Li2025LIPS,Li2025IneqSearch}. APPIRL learns certificate-basis choices, and NeuralSOS predicts monomial supports for an SOS optimizer \citep{Liu2025APPIRL,Pelleriti2026NeuralSOS}. AquaForte uses LLM-proposed function instantiations within a soundness-preserving SMT pipeline; its evaluation includes constructively generated problems asking whether a polynomial is a sum of exactly three squares \citep{Lv2026AquaForte}. Its target is quantified-SMT solving rather than sequential coordination of exact algebraic tools.

Among especially close recent systems, NSPI trains a direct SOS
conjecturer using synthetic polynomial--SOS data, SFT, and
curriculum-based reinforcement learning. It then applies numerical
refinement and rational recovery before Lean verification
\citep{Zuo2026NSPI}. We instead study sequential exact tool coordination
without numerical repair. The empirical object is therefore tool
coordination in a controlled certificate-search environment, not
state-of-the-art inequality proving.

\begin{table*}[t]
\centering
\small
\setlength{\tabcolsep}{5pt}
\begin{tabularx}{\textwidth}{p{0.15\textwidth}p{0.23\textwidth}p{0.29\textwidth}X}
\toprule
System & Learned role & Downstream symbolic component & Primary empirical object \\
\midrule
AIPS / LIPS / IneqSearch \citep{Wei2024AIPS,Li2025LIPS,Li2025IneqSearch} & Learned guidance, rewriting, ranking, or theorem selection & Symbolic transformations and inequality-proof search & Olympiad-style algebraic inequalities \\
\midrule
APPIRL \citep{Liu2025APPIRL} & Reinforcement learning for selecting certificate bases & Krivine-style polynomial certificate construction & Automated polynomial-inequality proofs \\
\midrule
NeuralSOS \citep{Pelleriti2026NeuralSOS} & Transformer prediction of an SOS monomial support/basis & SOS optimization and certificate checking & Learning-augmented nonnegativity certification \\
\midrule
NSPI \citep{Zuo2026NSPI} & LLM conjecture of an approximate SOS structure & Numerical refinement, rational recovery, and Lean checking & Formal multivariate inequality proving \\
\midrule
This work & Post-trained model--harness system coordinating native exact transformations & Exact SymPy observations and coefficient verifier; no numerical correction & Training and evaluating symbolic-tool coordination \\
\bottomrule
\end{tabularx}
\caption{Positioning relative to learning-based polynomial systems. The distinction is the scientific object: weighted SOS is used as an exactly checkable environment for studying a sequential tool agent rather than as a claim of general or state-of-the-art inequality solving.}
\label{tab:closest_work}
\end{table*}

\section{Verified Weighted-SOS Search}
\label{sec:task}

Let $R=\mathbb{Q}[x_1,\ldots,x_n]$ with a supplied variable order. Table~\ref{tab:task_suite} gives the complete curriculum and test interface. The eight supporting tasks are always solved directly, without native tool access, in both training and evaluation. Their purpose is algebraic supervision rather than a direct measurement of tool execution. Native function calling is used only for weighted-SOS search.

\begin{table*}[t]
\centering
\small
\setlength{\tabcolsep}{5pt}
\begin{tabularx}{\textwidth}{p{0.225\textwidth}p{0.205\textwidth}X}
\toprule
Task & Curriculum role & Output / evaluation \\
\midrule
\multicolumn{3}{l}{\emph{Tool-aligned local operations}}\\
Lexicographic term ordering
    & Local normalization
    & Equivalent polynomial in the requested order \\
Collection
    & Local normalization
    & Expression collected in the designated variable \\
Expansion
    & Local execution
    & Fully expanded polynomial \\
Factorization
    & Local structure exposure
    & Product representation \\
\midrule
\multicolumn{3}{l}{\emph{Broader structural algebra}}\\
Horner form
    & Structural representation
    & Nested form for a supplied variable order \\
Symmetric reduction
    & Structural representation
    & Expression in the designated symmetric basis \\
Quotient/remainder
    & Polynomial reduction
    & Exact $q,r$ with $f=qg+r$ and reduced $r$ \\
Extended GCD
    & Algebraic identity
    & Monic gcd with an exact B\'{e}zout identity \\
\midrule
\multicolumn{3}{l}{\emph{Composite target task}}\\
Weighted SOS
    & Certificate search
    & Native tools followed by exact terminal verification \\
\bottomrule
\end{tabularx}
\caption{Task taxonomy and curriculum roles. Dataset sizes and
variable/degree ranges are reported in Appendix; detailed output and acceptance
contracts are reported in the appendix.}
\label{tab:task_suite}
\end{table*}

\paragraph{Curriculum roles.}
The four local tasks mirror operations available through the deployment interface, but they are learned and evaluated without native calls. The four structural tasks require broader choices of representation---nested variable order, symmetric bases, reduced remainders, and B\'{e}zout identities---that are not returned by a single SOS tool invocation. Weighted SOS combines these capabilities into a sequential search problem. This grouping motivates the descriptive family analysis in the Results section. The grouping is descriptive, and per-task causal attribution is outside the experimental design.

\paragraph{Acceptance contract.}
A candidate is accepted only if it parses as a finite sum in the form of Eq.~\eqref{eq:intro-sos}, every weight is an exact positive rational number, every summand is a polynomial square, and
\[
\operatorname{expand}\!\left(f-\sum_j c_jp_j^2\right)=0
\]
coefficient by coefficient. Because every accepted summand is nonnegative on $\mathbb{R}^n$, this identity immediately certifies $f(a)\geq0$ for every real input $a$. Acceptance is not complete for polynomial nonnegativity, and the agent can also fail to find a certificate that exists. Every evaluation polynomial is SOS by construction, although the model is not told this; an exhausted or abstaining trajectory is therefore a failed search, not evidence that the polynomial is non-SOS.

\begin{figure*}[t]
\centering
\includegraphics[width=\textwidth]{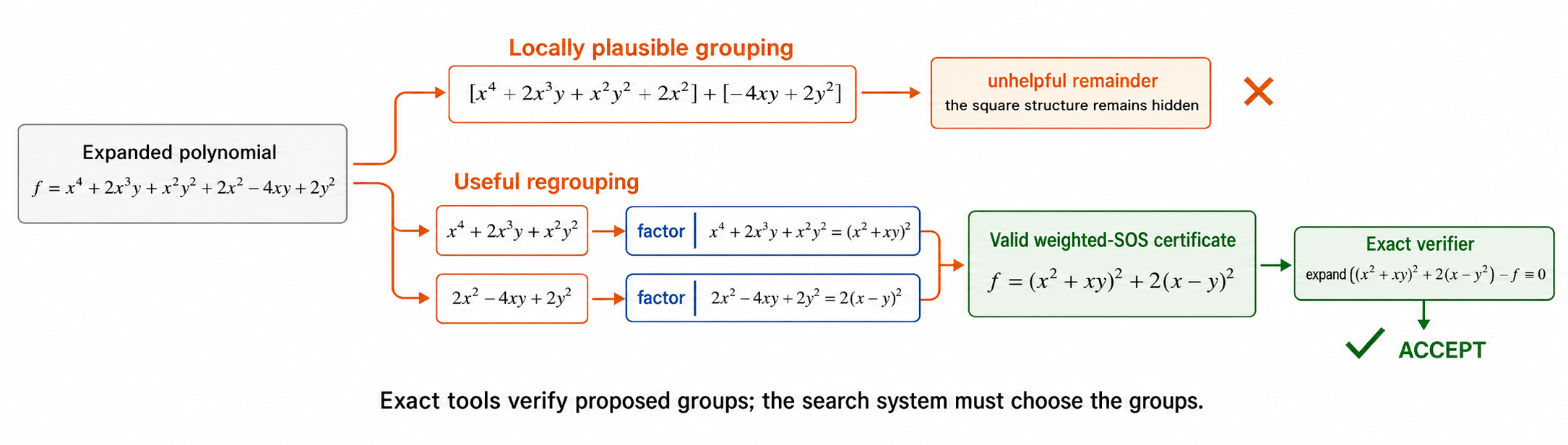}
\caption{Illustrative coordination example, not a model trajectory. Exact factorization verifies a proposed group, but the search system must decide which terms to group and when to revise an unhelpful partial decomposition. The useful regrouping exposes $(x^2+xy)^2+2(x-y)^2$, after which exact expansion verifies the certificate.}
\label{fig:coordination-example}
\end{figure*}

\section{Certificate Search as a Sequential Decision Problem}
\label{sec:agent-search}

\paragraph{Why coordination remains nontrivial.}
Figure~\ref{fig:coordination-example} separates local symbolic execution from global certificate search. A factorizer can verify either selected group exactly, but it does not choose the grouping. A locally plausible branch can therefore produce correct intermediate algebra and still leave a remainder that does not expose a weighted square. The model--harness system must revise that branch, assemble a complete candidate, and stop only when the terminal identity is ready for exact checking.

At the level of the complete model--harness system, weighted-SOS discovery is a finite-horizon sequential decision problem. At step $t$, the interaction history $h_t=(f,a_{<t},o_{<t},b_t)$
contains the target polynomial $f$, previous actions $a_{<t}$, exact symbolic observations $o_{<t}$, and the remaining call and attempt budget $b_t$. The next action may be a text continuation, a native symbolic request with concrete arguments, or a terminal certificate candidate. For an executable tool action, the backend deterministically returns an algebraic observation $o_t=T(a_t)$. For a terminal candidate $y$, success is determined by $r_{\mathrm{term}}(f,y)=\mathbf{1}_{\left[V(f,y)=1\right]},$
where $V$ is the exact structural and coefficient verifier defined above. The difficulty is therefore not uncertainty in local execution, but delayed credit assignment over a branching search: several algebraically correct calls may be strategically unhelpful, and success depends on the sequence of representation, regrouping, retry, and stopping decisions made within a finite budget.

This formulation motivates verifier-grounded policy optimization after SFT. Valid certificates and useful action sequences are non-unique, so imitating one reference trace would privilege one decomposition and one sequence of transformations while penalizing alternative valid solutions. Exact symbolic rewards instead allow multiple sampled responses to be compared by the algebraic state they reach rather than by surface agreement with a single trajectory. This is the role served by GRPO in our pipeline. The optimization and trajectory analysis are defined at the level of the complete model--harness system. Accordingly, trajectory-level results characterize the system as a whole rather than attributing every regrouping, retry, or stopping decision to the language model alone.

\paragraph{Native interface and system boundary.}
The weighted-SOS environment exposes \texttt{expand\_polynomial}, \texttt{collect\_terms}, \texttt{reorder\_polynomial}, and \texttt{factorize\_polynomial}. At each turn, the composite inference system emits text, a schema-checked native call, or a terminal object. The harness validates JSON, enforces budgets, executes valid calls exactly in SymPy, and appends the returned observation. Tool-enabled configurations share at most ten calls, at most three decomposition attempts, five seconds per call, and 60 seconds per episode. A terminal object is scored only by the exact certificate verifier. All trajectory-level results therefore characterize the complete model--harness system. The appendix summarizes this responsibility split and provides the full schemas.

\section{Algebra-Grounded Post-Training}
\label{sec:method}

\paragraph{Synthetic curriculum.}
Each of the nine tasks contributes $150{,}000$ generated examples, giving $1{,}350{,}000$ examples before a $5\%$ validation split and $1{,}282{,}500$ training records afterward. Inputs use one to three declared variables and one to five initial terms. Task-specific symbolic routines produce exact direct-task targets. Weighted-SOS instances are generated backward by sampling $m\in\{1,\ldots,5\}$, base polynomials $q_j$, and integer weights $c_j\in\{1,\ldots,10\}$, then displaying a scrambled expansion of $f=\operatorname{expand}\!\left(\sum_{j=1}^{m}c_jq_j^2\right).$
The sampled grouping is hidden. A separate seed generates $10{,}000$ test instances per task. Because train and test share generator families, the suite measures competence within the designed algebraic environment rather than natural-problem or out-of-distribution generalization.

\begin{table*}[t]
\centering
\small
\setlength{\tabcolsep}{4.5pt}
\begin{tabularx}{\textwidth}{p{0.14\textwidth}p{0.34\textwidth}p{0.20\textwidth}X}
\toprule
Stage & Model-facing representation & System interface & Learning/evaluation signal \\
\midrule
Direct-task SFT & ChatML problem and direct algebraic target & Direct text & Assistant-token cross-entropy \\
SOS SFT & Simulated Python-like symbolic snippets and later observed results & Simulated symbolic context; no native calls & Assistant-token cross-entropy; observations are context \\
\midrule
GRPO & SFT-initialized sampled responses serialized with the model chat template & Chat-template response optimization & Task-specific symbolic reward and format gate \\
\midrule
Direct evaluation & Final transformed expression & Direct text; no tools & Exact task contract \\
SOS evaluation & Native function calls, exact tool observations, and terminal object & Native calls for Base+Tools and Full & Exact weighted-SOS structural and coefficient verifier \\
\bottomrule
\end{tabularx}
\caption{Training and evaluation representations. The SFT corpus contains no native function-call messages. GRPO is defined by chat-template response optimization under symbolic rewards; native-call behavior is evaluated in the weighted-SOS deployment environment.}
\label{tab:representations}
\end{table*}

\paragraph{Supervised fine-tuning.}
We apply four-bit QLoRA to \texttt{Phi-4-reasoning-plus}. System and user tokens are retained as context but masked from the loss; only assistant tokens are prediction targets. Direct tasks use problem--answer records. SOS records include simulated symbolic traces with Python-like snippets and returned algebraic results, including retries and errors. Crucially, these records contain no native function-call messages or tool descriptions. The run uses rank $32$, scale $16$, dropout $0.05$, a 2,048-token limit, learning rate $5\times10^{-5}$, one epoch, and one H100. It comprises 181,848 packed sequences, approximately $3.72\times10^8$ tokens, and 22,731 optimizer steps; the appendix reports the complete training configuration.

\paragraph{Verifier-grounded GRPO.}
The second stage starts from the SFT model and optimizes sampled responses with task-specific symbolic rewards. Let $E(y)$ parse the required answer wrapper and return $\bot$ if it is absent. For task $k$,
\begin{equation}
\label{eq:main-reward}
R_k(y)=
\begin{cases}
-1, & E(y)=\bot,\\
0.1I_{\mathrm{fmt}}(y)+0.9r_k(y), & \text{otherwise},
\end{cases}
\end{equation}
where $r_k\in[0,1]$ is computed from the exact task contract rather than string equality. For example, quotient/remainder is checked through $f=qg+r$ and a reduced remainder, while SOS reward is gated on parseable weighted-square structure and coefficient residual. GRPO uses a global prompt batch of 64, eight generations per prompt, 20,039 optimizer steps, and approximately 55 hours on four 80GB H100 GPUs. We report this stage as checkpoint-level policy optimization under the symbolic objective above. Native-call trajectories are evaluated in the weighted-SOS deployment environment, and trajectory-level statements refer to the complete model--harness system.

\paragraph{Reward interpretation.}
The exact task scores reward the state produced by a response rather than surface agreement with one reference string. Local tasks combine algebraic equivalence with their requested normal form; structural tasks additionally check basis restrictions, remainder reduction, or B\'{e}zout identities; SOS reward is gated on positive-rational weighted-square structure and an exact coefficient residual. A separate trajectory-shaping term favors valid, executable, nonrepeated calls while assigning the largest increment to terminal verification. The appendix gives the full formulas. The study evaluates the combined reward and does not isolate individual reward components.

\paragraph{Native SOS deployment.}
At test time, Base+Tools and Full receive the native SymPy interface; Base and SFT do not. All eight supporting tasks remain direct prediction tasks for every checkpoint. Thus, the deployment-time mismatch is deliberate: SFT exposes the algebra and simulated effects of operations, whereas the final SOS system must interact through native calls and exact observations. The appendix provides the corresponding message formats, native schemas, responsibility split, and training settings. The current configurations establish the behavior of the complete deployed system, not the separate contribution of this representation change.

\section{Experiments}
\label{sec:experiments}

Our evaluation separates checkpoint-level algebraic competence from complete tool-enabled certificate search. We ask three questions: (1) how performance differs across Base, SFT, and Full on the eight direct polynomial tasks; (2) how Base, SFT, Base+Tools, and Full compare on exactly verified weighted-SOS search, where native tools are available only to the two tool-enabled systems; and (3) whether the specialized checkpoints show an obvious loss on two general-mathematics retention checks. We first define the evaluated systems and shared protocol, and then report supporting-task accuracy, weighted-SOS success, post-hoc task-family aggregates, terminal failure categories, and secondary retention results.

\begin{table}[t]
\centering
\small
\setlength{\tabcolsep}{4pt}
\begin{tabular}{lcc}
\toprule
Configuration & Checkpoint & Native tools on SOS \\
\midrule
Base & Unadapted Phi-4 & No \\
\addlinespace[2pt]
SFT & QLoRA SFT & No \\
\addlinespace[2pt]
Base+Tools & Unadapted Phi-4 & Yes \\
\addlinespace[2pt]
Full & SFT $\rightarrow$ GRPO & Yes \\
\bottomrule
\end{tabular}
\caption{Evaluated systems. Native tools are disabled on the eight supporting tasks and retention checks for all rows. Base+Tools and Full receive the same SOS interface and limits.}
\label{tab:configurations}
\end{table}

\paragraph{Configurations and fairness.}
Table~\ref{tab:configurations} separates checkpoint adaptation from SOS tool access. The same test instances, decoding settings, exact verifier, and applicable time/call budgets are held fixed across comparable conditions. This design compares complete systems. It does not isolate GRPO, adding tools after SFT, or their interaction.

\paragraph{Backbone scope.}
We use \texttt{Phi-4-reasoning-plus} throughout to hold pretraining and model capacity fixed. The software interfaces are model-agnostic, but empirical transfer to other backbones is not evaluated.

\paragraph{Evaluation layers.}
The supporting-task rows evaluate checkpoint skill under direct exact contracts, with native tools disabled. The weighted-SOS row evaluates complete deployed systems in a shared search environment, including the native interface for Base+Tools and Full. The terminal verifier then evaluates the mathematical state reached by that interaction. Keeping these layers separate avoids attributing direct-task checkpoint gains to unavailable tools or treating a syntactically valid call as a verified proof.

\paragraph{Benchmarks.}
The custom suite contains $10{,}000$ separately generated instances for each task, or $90{,}000$ total. We report exact accuracy; weighted-SOS success requires verifier acceptance. As secondary retention checks, we use the 90-problem AI-MO AIME validation set (AIME 2022--2024) and all 1,319 GSM-8K test problems \citep{AIMOAIME,Cobbe2021GSM8K}. Tools are disabled for these checks. Each checkpoint produces 16 samples per problem with temperature $0.65$, top-$p=0.95$, a 2,048-token limit, and seed 1234. \texttt{avg@16} is mean sample correctness, not pass@16 or majority vote.

\section{Results}
\label{sec:results}

We organize the results by evaluation layer. The eight supporting tasks measure checkpoint-level algebraic skill with native tools disabled. Weighted-SOS success measures the terminal state reached by the complete model--harness system, with native tools available only to Base+Tools and Full. We then use a post-hoc family aggregation to describe where the largest performance differences occur, examine how exact verification separates successful search from terminal failure, and conclude with general-mathematics retention checks and training cost.

\begin{table}[t]
\centering
\small
\setlength{\tabcolsep}{4pt}
\begin{tabular}{lrrr}
\toprule
Supporting task & Base & SFT & Full \\
\midrule
Term ordering & 76.34 & 91.28 & \textbf{97.61} \\
Collection & 58.71 & 84.36 & \textbf{96.84} \\
Expansion & 66.23 & 88.52 & \textbf{98.73} \\
Horner form & 33.87 & 72.14 & \textbf{89.47} \\
Symmetric reduction & 14.52 & 63.71 & \textbf{82.19} \\
Quotient/remainder & 41.16 & 79.83 & \textbf{93.28} \\
Extended GCD & 13.79 & 68.47 & \textbf{91.53} \\
Factorization & 31.44 & 76.29 & \textbf{97.12} \\
\midrule
Eight-task macro & 42.01 & 78.08 & \textbf{93.35} \\
\bottomrule
\end{tabular}
\caption{Accuracy (\%) on the eight supporting tasks. Native tools are disabled for every checkpoint on all rows, and each task contains $10{,}000$ test instances. The final row averages only these eight direct tasks.}
\label{tab:supporting_results}
\end{table}

\paragraph{Direct algebraic performance.}
Table~\ref{tab:supporting_results} shows that Full is strongest on every supporting task. Its eight-task macro-average is $93.35\%$, compared with $78.08\%$ for SFT and $42.01\%$ for Base. Because native tools are disabled for all checkpoints on these tasks, the rows measure checkpoint-level algebraic performance rather than deployment-time tool execution. The experiment reports checkpoint-level outcomes; attribution to particular training examples or reward components is outside its design.

\begin{table}[t]
\centering
\small
\setlength{\tabcolsep}{3.5pt}
\begin{tabular}{lrrrr}
\toprule
Metric & Base & SFT & B+T & Full \\
\midrule
Weighted SOS & 7.63 & 51.82 & 44.73 & \textbf{78.96} \\
Nine-task system macro & 38.19 & 75.16 & -- & \textbf{91.75} \\
\midrule
\multicolumn{5}{l}{\emph{General-mathematics retention checks}} \\
AIME-90, \texttt{avg@16} & 16.67 & 34.44 & -- & \textbf{38.89} \\
GSM-8K, \texttt{avg@16} & 91.58 & 93.71 & -- & \textbf{96.29} \\
\bottomrule
\end{tabular}
\caption{Weighted-SOS success, the system-level nine-task macro, and general-mathematics retention checks (\%). The nine-task macro averages the eight direct tasks in Table~\ref{tab:supporting_results} together with weighted SOS. B+T denotes Base+Tools and is reported only on weighted SOS, the task on which it differs operationally from Base. Native tools are disabled for both retention checks.}
\label{tab:sos_retention_results}
\end{table}

\paragraph{Verified weighted-SOS performance.}
Table~\ref{tab:sos_retention_results} gives the four-way comparison on weighted SOS, the only native tool-calling task. Full reaches $78.96\%$, compared with $51.82\%$ for SFT, $44.73\%$ for Base+Tools, and $7.63\%$ for Base. Relative to Base, SFT improves by $44.19$ percentage points and Base+Tools by $37.10$ points; Full is a further $27.14$ points above SFT and $34.23$ points above Base+Tools. These differences compare complete systems rather than isolated component effects.

Including weighted SOS yields system-level nine-task macro-averages of $38.19\%$, $75.16\%$, and $91.75\%$ for Base, SFT, and Full, respectively. Because the weighted-SOS component is evaluated with native tools only for the tool-enabled systems, this quantity combines eight direct checkpoint evaluations with one system-level certificate-search evaluation. It is therefore a heterogeneous summary of complete systems rather than a uniform checkpoint-only metric. Wilson intervals in the appendix quantify finite test-set sampling; the study does not estimate checkpoint-to-checkpoint variation.

\paragraph{Performance across algebraic task families.}
Table~\ref{tab:family_results} groups the reported tasks into a post-hoc descriptive taxonomy. Relative to SFT, Full gains $12.46$ points on local operations, $18.08$ points on broader structural algebra tasks, and $27.14$ points on weighted SOS. The first two differences compare checkpoints under tool-free direct evaluation, whereas the weighted-SOS difference compares complete deployed systems because Full additionally receives the native interface. The increasing gap is consistent with the complete system providing larger improvements as tasks require more global representation and search choices, but the grouping is descriptive rather than a component or per-task ablation.

\begin{table}[t]
\centering
\small
\setlength{\tabcolsep}{4.5pt}
\begin{tabular}{lrrrr}
\toprule
Post-hoc family & Base & SFT & Full & $\Delta$ \\
\midrule
Local operations & 58.18 & 85.11 & 97.58 & +12.46 \\
Structural algebra & 25.84 & 71.04 & 89.12 & +18.08 \\
Weighted SOS & 7.63 & 51.82 & 78.96 & +27.14 \\
\bottomrule
\end{tabular}
\caption{Descriptive family averages derived from Tables~\ref{tab:supporting_results} and~\ref{tab:sos_retention_results}; $\Delta$ is Full minus SFT in percentage points. Local operations comprise ordering, collection, expansion, and factorization; structural algebra comprises Horner form, symmetric reduction, quotient/remainder, and extended GCD. The grouping was chosen after inspecting task structure and is not a pre-registered analysis.}
\label{tab:family_results}
\end{table}

\paragraph{Verification separates search from proof.}
The verifier accepts only a structurally valid weighted-SOS expression that expands exactly to the target polynomial. Successful Full episodes use $3.8$ native calls on average, below the ten-call limit. Unsuccessful terminal outputs fall into three distinct classes: premature completion or abstention, an expression that is not a valid weighted SOS, and a well-formed weighted SOS whose expansion is the wrong polynomial. Exact verification prevents all three from being counted as proofs.

\paragraph{General-mathematics retention checks.}
Under one 16-sample evaluation pass, Full has the highest mean sample accuracy on AIME-90 ($38.89\%$) and GSM-8K ($96.29\%$). These secondary results are retention checks: they show no obvious loss of general mathematical performance in this evaluation pass, but they are not evidence of broad transfer or generalization. The comparison uses one backbone, one checkpoint per training stage, and no repeated-run uncertainty estimate.

\paragraph{Training budget.}
The SFT run uses one H100 for one epoch, corresponding to $22{,}731$ optimizer steps. The GRPO run uses $20{,}039$ optimizer steps and approximately 55 hours on four 80GB H100 GPUs. The appendix consolidates the training configuration.

\section{Discussion}

\paragraph{Exact execution does not remove coordination errors.}
A backend can execute a requested operation perfectly while the agent chooses an unhelpful operation, argument, grouping, or stopping point. The $44.73\%$ Base+Tools result and the three terminal failure types make this distinction concrete. Tool-agent evaluation should therefore inspect both the interaction and the mathematical state reached at the end, paralleling environment-state evaluations in broader tool-agent benchmarks \citep{Patil2025BFCL,Yao2025TauBench}.

\paragraph{Algebraic grounding is a plausible complement to tool access.}
The Full system's advantage over Base+Tools is consistent with the hypothesis that training on the operations and representations behind a tool can complement learning its interface. The direct-task rows and weighted-SOS row measure different objects: checkpoint skill without tools versus complete deployed search with tools. Causal effects of SFT, GRPO, individual supporting tasks, and the representation change from simulated traces to native calls are outside the four-configuration design.

\paragraph{Discovery, interaction, and verification must be distinguished.}
Every evaluation polynomial has a weighted-SOS certificate by construction, but the prompt does not reveal this fact and credit is awarded only for an accepted certificate. Thus the reported success rate is certificate-search yield under a fixed budget, not SOS recognition; abstention is a failed search, not a correct negative. Likewise, a schema-valid call and a successful backend execution are intermediate events rather than proofs. The terminal check is exact but narrower than proof-assistant verification: it certifies positive rational weights, square structure, and coefficient identity, without formalizing side assumptions or constrained Positivstellensatz arguments.

\section{Conclusion and Limitations}

Weighted-SOS search provides a controlled setting in which local algebra is exact, global search remains nontrivial, and final correctness is machine checkable. An algebra-grounded Phi-4 checkpoint trained with SFT and verifier-grounded GRPO, then deployed with native SymPy calls, reaches $78.96\%$ verified SOS success versus $44.73\%$ for the same base backbone with the same interface. Together with the direct-task results and terminal failure separation, this supports treating learned algebraic skill, tool interaction, and exact verification as distinct parts of a complete agent.

The evidence is limited to same-generator synthetic problems with one to three variables, weighted SOS, one backbone, and one run per stage. The experimental design compares complete configurations; component-level effects of SFT, GRPO, tools, individual supporting tasks, and the internal model--harness control split are outside its scope. The study does not cover natural inequality benchmarks, larger polynomials, non-SOS classification, constrained certificates, repeated training seeds, or cross-backbone transfer. These are substantive extensions rather than claims of the present work.

\section*{Acknowledgments} 

The authors acknowledge the National Artificial Intelligence Research Resource (NAIRR) Pilot and Voltage Park for providing the H100 GPU compute resources used in this research under allocation NAIRR250114.

\bibliography{references}

\clearpage
\appendix

\setcounter{figure}{0}
\renewcommand{\thefigure}{A\arabic{figure}}
\setcounter{table}{0}
\renewcommand{\thetable}{A\arabic{table}}
\setcounter{equation}{0}
\renewcommand{\theequation}{A\arabic{equation}}

\section{Detailed Task Contracts and Exact Verifiers}
\label{app:contracts}

This section expands the compact task summary in Table~\ref{tab:task_suite}. All supporting tasks are direct prediction tasks; only weighted-SOS evaluation exposes the native SymPy interface.

\begin{table*}[t]
\centering
\small
\renewcommand{\arraystretch}{1.08}
\setlength{\tabcolsep}{5.5pt}
\begin{tabularx}{\textwidth}{p{0.18\textwidth}p{0.29\textwidth}X p{0.11\textwidth}}
\toprule
Task & Required output & Exact acceptance or reward check & Native tools \\
\midrule
\multicolumn{4}{l}{\emph{Tool-aligned local operations}}\\
Term ordering & Equivalent polynomial in a requested lexicographic monomial order & Polynomial equality plus requested order/normal-form check & No \\
Collection & Expression collected with respect to a selected variable or expression & Exact polynomial equality and coefficient-dictionary agreement & No \\
Expansion & Fully expanded polynomial & Exact equality and expanded-normal-form check & No \\
Factorization & Product of factors and multiplicities & Exact expanded product; reward additionally checks irreducibility over the supported domain & No \\
\midrule
\multicolumn{4}{l}{\emph{Broader structural algebra}}\\
Horner form & Nested expression for a supplied variable order & Exact equality and recursive Horner-structure depth & No \\
Symmetric reduction & Expression in a designated elementary-symmetric basis & Substitute the basis polynomials and compare exactly with the target & No \\
Quotient/remainder & $q,r$ such that $f=qg+r$ & Exact identity plus no remainder monomial divisible by $\operatorname{LT}(g)$ & No \\
Extended GCD & Monic $d$ and B\'{e}zout coefficients $a,b$ & $d\mid f$, $d\mid g$, $af+bg=d$, and monic normalization in $\mathbb{Q}[x]$ & No \\
\midrule
\multicolumn{4}{l}{\emph{Composite target task}}\\
Weighted SOS & Wrapped status and $\sum_jc_jp_j^2$ & Positive rational weights, polynomial-square structure, and exact coefficient identity with $f$ & Yes, for Base+Tools and Full \\
\bottomrule
\end{tabularx}
\caption{Detailed task contracts. Exact symbolic equality is evaluated with SymPy. Horizontal group rules separate local operations, broader structural tasks, and the composite certificate task.}
\label{tab:contracts}
\end{table*}

\subsection{Weighted-SOS acceptance}

A submitted certificate must parse as
\begin{equation}
\label{eq:app-weighted-sos}
f=\sum_{j=1}^{m}c_jp_j^2,
\qquad c_j\in\mathbb{Q}_{>0},\quad p_j\in R.
\end{equation}
The verifier first checks the structure of every summand, then computes
\[
\operatorname{expand}\!\left(f-\sum_{j=1}^{m}c_jp_j^2\right)
\]
and requires every coefficient to be exactly zero. It accepts any valid decomposition, not only the generator's hidden one. A syntactically valid answer can therefore fail either structurally or by identity.

\subsection{Terminal failure taxonomy}

\begin{figure*}[!t]
\centering
\includegraphics[width=\textwidth]{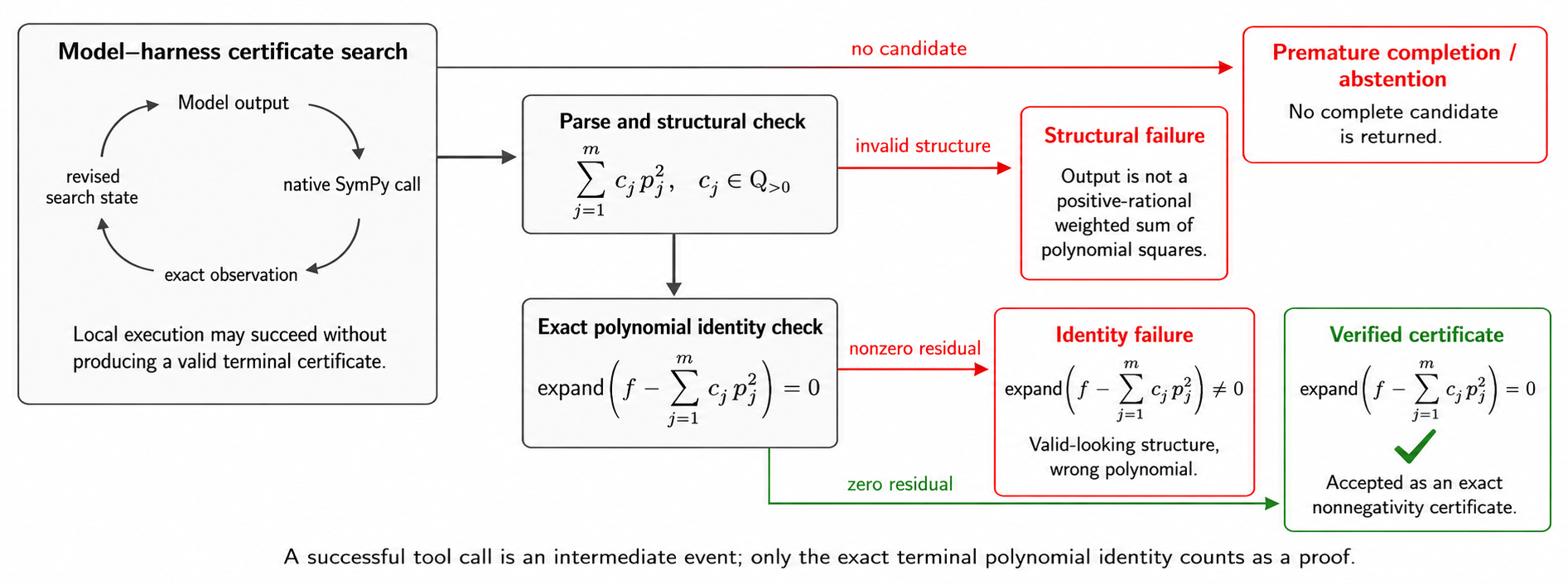}
\caption{Terminal verification taxonomy. Tool execution and schema validity are intermediate properties. The exact verifier separately detects premature completion, invalid weighted-SOS structure, and coefficient-identity failure, accepting only a positive-rational weighted-SOS certificate whose expansion equals the input polynomial exactly.}
\label{fig:failure-modes}
\end{figure*}

Figure~\ref{fig:failure-modes} separates interface events from mathematical acceptance. A schema-valid call or successful symbolic execution can still lead to a terminal failure; only the final coefficient identity constitutes a certificate.

An \emph{invalid call} uses an unregistered function or an argument object that fails schema validation. An \emph{execution error} includes parsing failure, timeout, or a symbolic-backend exception. A \emph{repeated call} is an identical call repeated within a trajectory. Terminal mathematical failures are separated as follows:
\begin{enumerate}
    \item \textbf{Premature completion or abstention:} the system stops without returning an accepted certificate.
    \item \textbf{Structural failure:} the returned expression cannot be parsed as a weighted sum of polynomial squares with exact positive rational weights.
    \item \textbf{Identity failure:} the expression has valid weighted-square structure but does not expand exactly to the target polynomial.
\end{enumerate}
The exact verifier accepts $78.96\%$ of Full SOS episodes and rejects every terminal output in these three classes.

\subsection{General constrained context outside the evaluation scope}
For inequalities $g_i(x)\geq0$ defining a semialgebraic set $K$, a preordering-style certificate can take the form \begin{equation} \label{eq:app-constrained} f=\sum_{S\in\mathcal A}\sigma_Sg_S, \qquad \sigma_S=\sum_jc_{Sj}p_{Sj}^2, \end{equation} where $g_S=\prod_{i\in S}g_i$ and $c_{Sj}>0$. When $\mathcal A=\mathcal P(\{1,\ldots,r\})$, this is a preordering certificate. Schm\"udgen's theorem guarantees such a representation for polynomials strictly positive on a compact basic closed set \citep{Schmuedgen1991KMoment}. Under an Archimedean quadratic-module hypothesis, Putinar instead gives the smaller form $f=\sigma_0+\sum_i\sigma_i g_i$ \citep{Putinar1993PositivePolynomials}. Stengle's general Positivstellensatz may require additional multiplier, denominator, or ideal terms \citep{Stengle1974}, while SOS constraints can be searched through semidefinite relaxations \citep{Parrilo2003SDPRelaxations}. The main experiments set $\mathcal A=\{\varnothing\}$; constrained-certificate competence is outside the evaluation scope.

\section{Synthetic Data Generation}
\label{app:data}

\subsection{Shared sampler and construction principles}
\label{app:shared-generator}

The nine datasets share a compact polynomial sampler and differ in how the sampled objects are transformed into a problem--target pair. A generic sampled polynomial has the form
\begin{equation}
\label{eq:shared-polynomial-sampler}
p(x_1,\ldots,x_n)=\sum_{\ell=1}^{L}a_\ell
\prod_{i=1}^{n}x_i^{\alpha_{\ell i}},
\end{equation}
with one to three declared variables, one to five initial terms, and nonzero integer coefficients $a_\ell\in[-10,10]$. Duplicate monomials are combined symbolically, and task adapters control exponent ranges so that the realized degrees match Table~\ref{tab:dataset_summary}. The displayed variable priority is sampled independently, so the same algebraic object can be requested under different lexicographic orders. Training and evaluation generation use seeds 42 and 2025, respectively.

The supporting tasks use two complementary construction patterns. \emph{Forward transformations} begin with a sampled expression and apply one deterministic symbolic operation to obtain the target. Ordering, collection, expansion, Horner form, and symmetric reduction follow this pattern. \emph{Algebraic-relation tasks} require several outputs to satisfy a common identity. Quotient/remainder and extended GCD are generated and solved under exact polynomial arithmetic, while factorization is constructed backward from sampled factors. In every case, the requested answer is specified by an invariant---polynomial equality, a normal-form condition, a reduced remainder, or a B\'{e}zout identity---rather than by one surface string.

\begin{table}[!t]
\centering
\small
\renewcommand{\arraystretch}{1.08}
\setlength{\tabcolsep}{4.5pt}
\begin{tabularx}{\columnwidth}{@{}p{0.46\columnwidth}cX@{}}
\toprule
Task & Variables & Realized degree \\
\midrule
Term ordering & 1--3 & 0--10 \\
Collection & 1--3 & 0--10 \\
Expansion & 1--3 & 0--55 \\
Factorization & 1--3 & 2--6 \\
Horner form & 1--3 & 0--20 \\
Symmetric reduction & 1--3 & 0--5 \\
Quotient/remainder & 1--3 & $f$: 1--6; $g$: 1--3 \\
Extended GCD & 1 & 2--10 \\
Weighted SOS & 1--3 & 2, 4, or 6 \\
\bottomrule
\end{tabularx}
\caption{Generator ranges. Each task contains 150,000 generated records before the common $5\%$ validation split and 10,000 separately generated evaluation instances, for 1.35 million records and 90,000 evaluation instances in total. Algebraic composition can produce coefficients outside the atomic sampling range.}
\label{tab:dataset_summary}
\end{table}

\subsection{Weighted-SOS backward construction}
\label{app:sos-generation}

\begin{figure*}[!t]
\centering
\includegraphics[width=\textwidth]{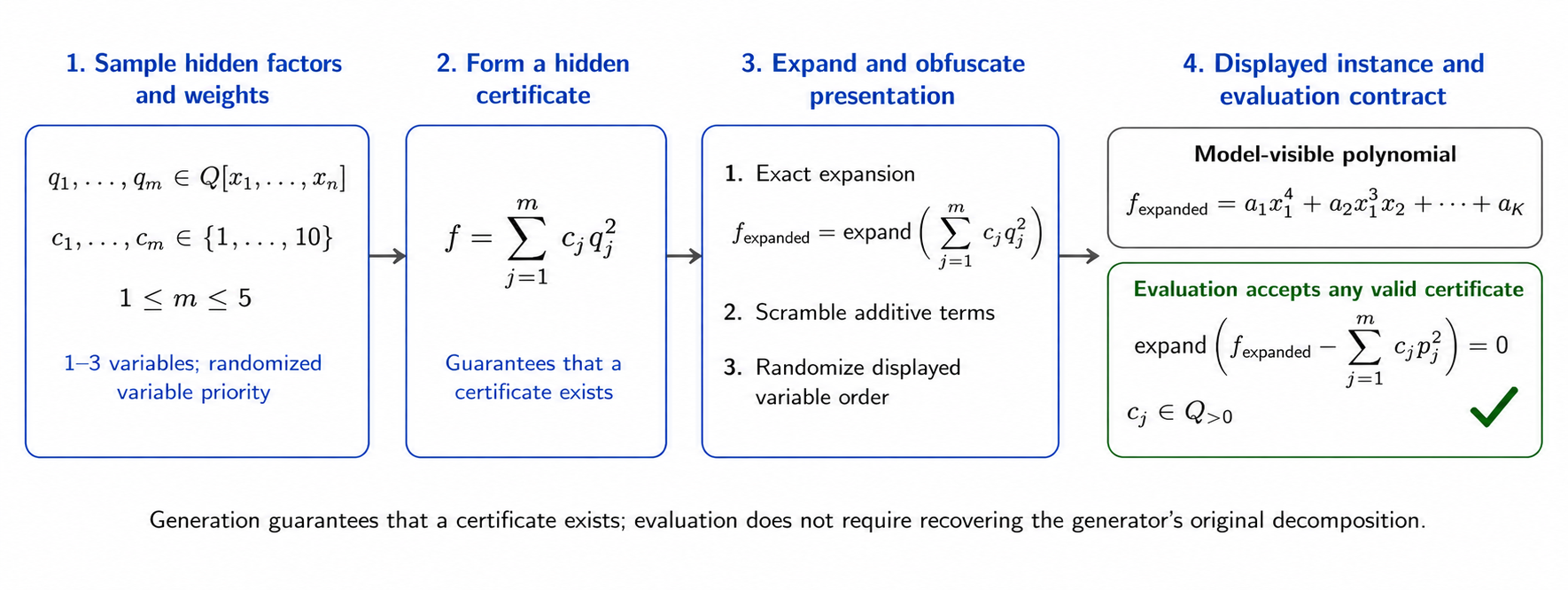}
\caption{Weighted-SOS instance generation. The generator samples a hidden positive-weight certificate, expands it exactly, and randomizes the displayed presentation. The hidden decomposition guarantees existence but is not shown to the model, and evaluation accepts any certificate satisfying the exact polynomial identity.}
\label{fig:data-generation}
\end{figure*}

The composite generator samples $m\in\{1,\ldots,5\}$, base polynomials $q_1,\ldots,q_m$, and integer weights $c_j\in\{1,\ldots,10\}$, and computes
\begin{equation}
\label{eq:sos-generation}
f=\operatorname{expand}\!\left(\sum_{j=1}^{m}c_jq_j^2\right).
\end{equation}
The expanded terms and displayed variable order are randomized before the problem is serialized. The sampled decomposition guarantees existence but is never required as the answer: any positive-rational weighted-SOS expression satisfying the exact identity is accepted. Figure~\ref{fig:data-generation} summarizes this backward-construction pipeline.

\subsection{Normal forms and local transformations}
\label{app:local-task-generation}

\paragraph{Lexicographic term ordering.}
The generator samples an expanded polynomial, randomizes its displayed additive order, and supplies a variable precedence $x_{\pi(1)}\gg\cdots\gg x_{\pi(n)}$. Every monomial is represented by the exponent tuple
\[
(\alpha_{\ell,\pi(1)},\ldots,\alpha_{\ell,\pi(n)}),
\]
and the target lists terms in descending lexicographic order of these tuples. Coefficients do not affect the ordering except through cancellation when duplicate monomials are combined. For example, under $y\gg x\gg z$,
\[
3x^2yz-4y^3x+2z^4+5x^2y^2
\]
is serialized as
\[
-4y^3x+5y^2x^2+3yx^2z+2z^4.
\]
The verifier checks both polynomial equality and the emitted monomial sequence.

\paragraph{Collection.}
A collection instance pairs a sampled polynomial with a designated variable $v$. After exact expansion and combination of like terms, the target is the coefficient decomposition
\begin{equation}
\label{eq:collection-form}
p=\sum_{j=0}^{d_v}C_j(x_1,\ldots,\widehat{v},\ldots,x_n)v^j,
\end{equation}
where the hat denotes omission of $v$ and each $C_j$ is ordered under the remaining variable precedence. For instance,
\[
2x^2y+x^2z+3xy^2+4xz+7
=(2y+z)x^2+(3y^2+4z)x+7.
\]
This task therefore tests more than equality: the answer must expose the requested coefficient structure rather than return a flat expansion.

\paragraph{Expansion.}
Expansion instances begin with compact expressions containing sums, products, and positive integer powers. The exact target is obtained by recursively distributing products over sums, expanding powers, combining equal monomials, and finally applying the requested term order. A representative source expression is
\[
(x+yz)(2x-y^2+z),
\]
whose expanded value is
\[
2x^2-xy^2+xz+2xyz-y^3z+yz^2.
\]
Unlike collection, expansion removes the expression tree introduced by products and powers. This distinction exposes the model to two operationally different normalizations: one flattens an expression, while the other organizes a flat polynomial around a selected variable.

\subsection{Recursive and invariant representations}
\label{app:structural-task-generation}

\paragraph{Horner form.}
The Horner adapter supplies an expanded polynomial and an ordered list of variables. For the first variable $v$, it computes the coefficients in Eq.~\eqref{eq:collection-form}, inserts zero coefficients for missing powers, and forms the nested expression
\begin{equation}
\label{eq:horner-recursion}
H_v(p)=((C_dv+C_{d-1})v+\cdots+C_1)v+C_0.
\end{equation}
For multiple variables, the same construction is applied recursively to the coefficient polynomials using the remaining supplied order. For example,
\begin{align*}
p={}&(3y^2z+2yz)x^2+(-5y^3z+y^2+4z)x\\
&{}-2y^2z+3
\end{align*}
becomes
\[
\bigl((3y^2z+2yz)x-5y^3z+y^2+4z\bigr)x-2y^2z+3.
\]
The acceptance check expands the answer exactly and also measures whether the requested nesting is present at each recursive level.

\paragraph{Symmetric reduction.}
These instances use a designated variable set and the corresponding elementary symmetric polynomials
\[
e_k(x_1,\ldots,x_n)=
\sum_{1\le i_1<\cdots<i_k\le n}x_{i_1}\cdots x_{i_k}.
\]
For a supported symmetric input, the target is a polynomial $Q$ satisfying
\begin{equation}
\label{eq:symmetric-target}
p(x_1,\ldots,x_n)=Q(e_1,\ldots,e_n).
\end{equation}
The exact reduction eliminates the leading monomial iteratively. If $\operatorname{LM}(p)=x_1^{a_1}\cdots x_n^{a_n}$ with $a_1\ge\cdots\ge a_n$, then
\[
e_1^{a_1-a_2}e_2^{a_2-a_3}\cdots e_n^{a_n}
\]
has the same leading monomial under the designated lexicographic order; subtracting its suitable multiple strictly lowers the leading term. The task uses the zero-remainder case and serializes only $Q$. For two variables,
\[
x^3+y^3=e_1^3-3e_1e_2,
\qquad e_1=x+y,
\quad e_2=xy.
\]
Evaluation substitutes the elementary symmetric polynomials back into the proposed expression and compares the resulting polynomial with the input exactly.

\subsection{Algebraic relations and backward construction}
\label{app:relation-task-generation}

\paragraph{Quotient and remainder.}
The adapter samples four polynomials $p_0,p_1,p_2,p_3$, forms
\begin{equation}
\label{eq:division-generation}
f=p_0p_1+p_2,
\qquad g=p_3\ne0,
\end{equation}
and then runs exact single-divisor multivariate division under the requested monomial order. The product-plus-offset construction produces algebraic dependence without assuming that the final quotient is $p_0$ or that the final remainder is $p_2$; the canonical pair is determined by repeated leading-term reduction. At each step, if $\operatorname{LM}(g)$ divides the current leading monomial, the appropriate multiple of $g$ is subtracted and added to $q$; otherwise the current leading term is moved to $r$. The process terminates with
\[
f=qg+r,
\qquad
\operatorname{LM}(g)\nmid m
\quad\text{for every monomial }m\text{ of }r.
\]
For example, under lexicographic order $x\gg y$,
\[
f=x^2+xy,
\qquad g=xy-1
\]
yields $q=1$ and $r=x^2+1$.

\paragraph{Extended GCD.}
Extended-GCD instances are univariate, so the computation remains in the Euclidean domain $\mathbb{Q}[x]$. Given sampled nonzero polynomials $f$ and $g$, exact polynomial division generates the remainder sequence
\[
r_{i-1}=q_ir_i+r_{i+1},
\]
while the same updates are applied to coefficient pairs expressing each $r_i$ as a linear combination of the original inputs. When the remainder becomes zero, the last nonzero remainder is normalized to a monic polynomial $d$, and the corresponding coefficients $a,b$ satisfy
\begin{equation}
\label{eq:bezout-target}
af+bg=d=\gcd(f,g).
\end{equation}
For $f=x^3-1$ and $g=x^2-1$, one valid target is
\[
d=x-1,
\qquad a=1,
\qquad b=-x,
\]
since $(x^3-1)-x(x^2-1)=x-1$. The verifier checks divisibility, the B\'{e}zout identity, and monic normalization rather than one particular coefficient pair.

\paragraph{Factorization.}
Factorization is generated backward. The adapter samples an integer content and nonconstant polynomial factors, multiplies them, and expands the product before presenting it as the input. The exact target records the content and factor--multiplicity pairs over the supported coefficient domain. For example,
\[
6(x+2y)(xy+1)=6x^2y+12xy^2+6x+12y.
\]
The prompt contains the expanded right-hand side, while the target exposes the product on the left. Because units, factor order, and repeated-factor grouping admit equivalent presentations, the checker expands the proposed product exactly; the graded reward additionally measures whether the reported nonconstant factors are irreducible over the supported domain.

\paragraph{Relation to the composite task.}
The eight generators deliberately separate operations that are entangled during SOS search. Ordering and collection control presentation; expansion checks a proposed identity; factorization exposes candidate square structure; Horner and symmetric reduction train non-flat representations; division and extended GCD enforce exact multi-object identities. Weighted-SOS search combines several of these capabilities but does not reduce to any one supporting task.

\section{Training--Deployment Boundary}
\label{app:train-deploy}

Supporting-task supervision and deployment-time tool access use different interfaces. SFT teaches algebraic transformations through direct prediction and simulated symbolic observations without native function-call messages. GRPO optimizes chat-template responses with task-specific symbolic rewards. During weighted-SOS evaluation, Base+Tools and Full use the native SymPy interface, while the eight supporting tasks and the retention checks remain direct, tool-free evaluations. Terminal success is determined independently by the exact certificate verifier.

\section{Native Tool and Output Schemas}
\label{app:tools}

\begin{center}
\begin{minipage}{\columnwidth}
\centering
\small
\renewcommand{\arraystretch}{1.08}
\setlength{\tabcolsep}{4.5pt}
\begin{tabularx}{\columnwidth}{@{}p{0.25\columnwidth}X@{}}
\toprule
Component & Responsibility \\
\midrule
Model-facing output & Produces a text continuation, a native call object, or a terminal answer object. \\
Harness & Checks call schemas, executes valid SymPy functions, returns observations, and enforces call, attempt, strike, and time limits. \\
Terminal verifier & Parses weighted squares, enforces positive rational weights, and compares the expanded coefficient dictionary exactly. \\
\bottomrule
\end{tabularx}
\captionof{table}{Responsibility split for weighted-SOS evaluation. Trajectory-level statistics characterize the complete model--harness system, while terminal correctness is determined independently by the exact verifier.}
\label{tab:responsibility}
\end{minipage}
\end{center}

\begin{center}
\begin{minipage}{\columnwidth}
\centering
\small
\renewcommand{\arraystretch}{1.08}
\setlength{\tabcolsep}{4pt}
\begin{tabularx}{\columnwidth}{@{}p{0.48\columnwidth}X@{}}
\toprule
Function & Argument and result contract \\
\midrule
\texttt{expand\_polynomial} & Required \texttt{expression}; returns \texttt{status}, \texttt{original}, and \texttt{expanded}, or an error message. \\
\texttt{collect\_terms} & Required \texttt{expression}, optional \texttt{variable}; returns \texttt{collected} with status fields. \\
\texttt{reorder\_polynomial} & Required \texttt{expression}, optional \texttt{order}; returns \texttt{reordered} and the applied order. \\
\texttt{factorize\_polynomial} & Required \texttt{expression}; returns \texttt{factorized} with status fields. \\
\bottomrule
\end{tabularx}
\captionof{table}{Native functions used by Base+Tools and Full during weighted-SOS evaluation.}
\label{tab:tool_schemas}
\end{minipage}
\end{center}

Every call contains an identifier, a registered function name, and a JSON argument object. Undeclared fields are rejected. Tool expressions use explicit \texttt{*} for multiplication and \texttt{**} for powers. Parse or backend exceptions return \texttt{status=error} with a message. A call has a five-second limit; an episode has a 60-second limit; three consecutive schema-invalid calls terminate the episode. The system permits at most ten calls and three decomposition attempts.

The terminal answer object contains the fields \texttt{success}, \texttt{is\_sos}, \texttt{attempts\_used}, \texttt{SOSPolynomial}, and \texttt{reason}. Because every evaluation input is SOS, a false \texttt{is\_sos} or unsuccessful status is treated as failure/abstention, not a verified mathematical classification.

\section{Message Formats}
\label{app:formats}

\paragraph{Direct-task SFT records.}
The corpus uses multi-turn ChatML messages. System and user messages state the task and inputs; assistant turns contain the target transformation or algebraic derivation inside task-specific wrappers. System/user tokens are visible as context but masked from the SFT loss.

\paragraph{Simulated symbolic SFT traces.}
SOS records represent symbolic computation through wrapped Python-like snippets followed by evaluated results in later context. The assistant is trained on its reasoning, snippets, and final answer; symbolic observations are context rather than prediction targets. These records are not native \texttt{tool\_calls}, and the corpus does not expose the native tool schemas.

\paragraph{GRPO prompts and outputs.}
GRPO inputs and outputs use the Phi-4 chat template and are
scored by the exact task-specific parsers and symbolic rewards
described in the main paper. We report GRPO as checkpoint-level response optimization. Native-interaction analysis concerns the weighted-SOS evaluation environment, and trajectory-level conclusions refer to the complete model--harness system.

\section{Training and Reproducibility Details}
\label{app:training}

Table~\ref{tab:training_hparams} consolidates the SFT and GRPO settings used in the completed runs.

\begin{table*}[!t]
\centering
\small
\renewcommand{\arraystretch}{1.10}
\setlength{\tabcolsep}{5pt}
\begin{tabularx}{\textwidth}{@{}p{0.17\textwidth}XX@{}}
\toprule
Setting & \textbf{Supervised fine-tuning} & \textbf{Group Relative Policy Optimization} \\
\midrule
Initialization & \texttt{Phi-4-reasoning-plus}; four-bit weights & SFT model and loaded adapter \\
Data and signal & Multi-turn ChatML; $5\%$ validation; assistant-token cross-entropy & Multi-task prompts; exact symbolic task rewards \\
Adaptation & LoRA rank 32, scale 16, dropout 0.05 & Continued adapter training; rank 32, scale 16, dropout 0.05 \\
Length and packing & 2,048 tokens with packing & \\
Length limits & & 2,048 prompt and 2,048 completion tokens \\
Horizon & One epoch; 22,731 optimizer steps & One epoch; 20,039 optimizer steps; approximately 55 hours \\
Batching & Micro-batch 1; accumulation 8 & Per-device batch 1; four processes; accumulation 16; eight generations per prompt \\
Optimization & LR $5\times10^{-5}$; cosine; 100 warmup steps; 8-bit AdamW & LR $10^{-6}$; cosine; 50 warmup steps \\
Precision/hardware & bfloat16; gradient checkpointing; one H100 & bfloat16; four 80GB H100s; ZeRO-2 CPU offload \\
Sampling and seed & Teacher forcing; seed 42 & Temperature 0.9; top-$p=0.95$; seed 42 \\
\bottomrule
\end{tabularx}
\caption{Training settings for the completed SFT and GRPO runs.}
\label{tab:training_hparams}
\end{table*}

The SFT environment uses Axolotl 0.8.0-dev, PEFT 0.14.0, Transformers 4.49.0, PyTorch $2.5.1+\mathrm{cu124}$, Datasets 3.2.0, and Tokenizers 0.21.0. The run contains 181,848 packed sequences, approximately $3.72\times10^8$ tokens, and 7.05 raw examples per packed sequence; its validation loss is 0.0158.

\paragraph{Backbone and portability.}
All reported checkpoints share the same Phi-4 backbone so that the study focuses on post-training and deployed coordination rather than model-family scaling. The generators, exact contracts, native schemas, SymPy backend, and terminal verifier are model-agnostic software components. Reusing them with another backbone is technically straightforward, but matched empirical transfer requires repeating both SFT and verifier-grounded post-training and is outside the present evidence.

\section{Reward Functions}
\label{app:rewards}

Let $E(y)$ parse the task-specific object between \texttt{<<<} and \texttt{>>>}, and let $E(y)=\bot$ when the wrapper is missing. Each parseable object receives a format indicator $I_{\mathrm{fmt}}$ and task score $r_k\in[0,1]$:
\begin{equation}
\label{eq:grpo_reward}
R_k(y)=
\begin{cases}
-1, & E(y)=\bot,\\
0.1 I_{\mathrm{fmt}}(y)+0.9r_k(y), & \text{otherwise}.
\end{cases}
\end{equation}
All symbolic equalities are evaluated exactly. Write $I[\cdot]$ for an indicator and $\mathbf c(e)$ for the coefficient vector of expanded polynomial $e$. The normalized residual score is
\begin{equation}
\label{eq:graded_residual}
\rho_f(e)=\exp\!\left(-\lambda\frac{\|\mathbf c(e)\|_1}{\|\mathbf c(f)\|_1+\varepsilon}\right),
\qquad \lambda=10.
\end{equation}
For collection, let $\mathcal I$ be the union of coefficient-dictionary supports of the target and output. The task scores are
\begin{align}
r_{\mathrm{ord}} &=0.3I[p=f]+0.7\tau(p), \\
r_{\mathrm{collect}} &=0.3I[p=f] \notag\\[-2pt]
&\quad+0.7\frac{|\{i\in\mathcal I:c_i^p=c_i^f\}|}{\max(1,|\mathcal I|)}, \\
r_{\mathrm{expand}} &=0.5I[p=f]+0.5I[p=\operatorname{expand}(p)], \\
r_{\mathrm{Horner}} &=0.4I[p=f] \notag\\[-2pt]
&\quad+0.6\frac{d_{\mathrm{correct}}}{\max(1,\deg f)}, \\
r_{\mathrm{sym}} &=0.7I[p(e_1,\ldots,e_n)=f] \notag\\[-2pt]
&\quad+0.3I[\operatorname{vars}(p)\subseteq\{e_i\}], \\
r_{\mathrm{div}} &=0.6\rho_f(f-qg-r) \notag\\[-2pt]
&\quad+0.4I[r\text{ is reduced by }\operatorname{LT}(g)], \\
r_{\mathrm{xgcd}} &=0.5\rho_f(af+bg-d) \notag\\[-2pt]
&\quad+0.3I[d\mid f\wedge d\mid g]+0.2I[d\text{ monic}], \\
r_{\mathrm{factor}} &=0.5I[\prod_jh_j^{e_j}=f] \notag\\[-2pt]
&\quad+0.5\frac{\sum_je_j\deg(h_j)I[h_j\text{ irreducible}]}{\deg f}, \\
r_{\mathrm{SOS}} &=0.3I_{\mathrm{struct}} \notag\\[-2pt]
&\quad+0.7I_{\mathrm{struct}}\rho_f\!\left(f-\sum_jc_jp_j^2\right).
\end{align}
Here $\tau\in[0,1]$ is normalized Kendall agreement with the requested monomial order and $d_{\mathrm{correct}}$ counts correct recursive Horner levels. The SOS structural gate requires exact positive rational weights and polynomial squares.

The tool-trajectory shaping reward is
\begin{equation}
\label{eq:tool_reward}
\begin{aligned}
R_{\mathrm{tool}}={}&0.05N_{\mathrm{schema}}+0.05N_{\mathrm{exec}}\\[-2pt]
&-0.2N_{\mathrm{repeat}}-0.01T+I_{\mathrm{verified}}.
\end{aligned}
\end{equation}
It rewards schema-valid calls and successful execution, penalizes repeated identical calls and long trajectories, and gives the largest single increment to a verified terminal certificate. The study evaluates this combined reward and does not isolate individual reward components.

\section{Evaluation Protocol}
\label{app:evaluation}

\paragraph{Custom tasks.}
All comparable configurations use the same $10{,}000$ items per task, decoding procedure, answer parser, exact verifier, and applicable computational budgets. Weighted-SOS Base+Tools and Full share the four native functions, ten-call limit, three-attempt limit, five-second call limit, and 60-second episode limit. The custom-task results are reported under this shared protocol.

\paragraph{General-mathematics retention checks.}
The evaluation uses all 90 examples in \texttt{AI-MO/aimo-validation-aime}, covering AIME 2022--2024, and all 1,319 GSM-8K test problems. Native tools are disabled. Each checkpoint generates 16 samples per problem with temperature 0.65, top-$p=0.95$, at most 2,048 output tokens, and seed 1234. The reported \texttt{avg@16} is the mean of the correctness indicator over all samples; it is neither pass@16 nor majority-vote accuracy.

\section{Additional Descriptive Results}
\label{app:additional-results}

The intervals below quantify uncertainty from the finite evaluation set under an independent Bernoulli-item interpretation. They do not estimate checkpoint-to-checkpoint or training-run variation.

\begin{center}
\begin{minipage}{\columnwidth}
\centering
\small
\renewcommand{\arraystretch}{1.08}
\setlength{\tabcolsep}{4.5pt}
\begin{tabular}{@{}lrr@{}}
\toprule
SOS configuration & Success & Wilson 95\% interval \\
\midrule
Base & 7.63\% & [7.13, 8.17] \\
SFT & 51.82\% & [50.84, 52.80] \\
Base+Tools & 44.73\% & [43.76, 45.71] \\
Full & 78.96\% & [78.15, 79.75] \\
\bottomrule
\end{tabular}
\captionof{table}{Approximate Wilson intervals treating the 10,000 test instances as independent Bernoulli trials. They quantify finite test-set sampling, not variation across checkpoints or training runs.}
\label{tab:app-ci}
\end{minipage}
\end{center}

\paragraph{Tool-use summary.}
Successful Full episodes use 3.8 native calls on average, measured over verifier-accepted weighted-SOS episodes.

\end{document}